\documentclass[runningheads]{llncs}
\usepackage[inkscapelatex=false]{svg}
\usepackage{multirow}
\usepackage{amsmath}
\usepackage{wrapfig}
\usepackage{tcolorbox}
\usepackage{booktabs}

\usepackage{fontawesome5}

\usepackage[T1]{fontenc}
\usepackage{graphicx}
\usepackage{hyperref}
\usepackage{color}

\begin{document}
\title{LITA: An Efficient LLM-assisted Iterative Topic Augmentation Framework}
%
%

\author{Chia-Hsuan Chang\inst{1}\orcidID{0000-0001-9116-8244}\faIcon{envelope} \and
Jui-Tse Tsai\inst{2} \and
Yi-Hang Tsai\inst{2} \and
San-Yih Hwang\inst{2}\faIcon{envelope}}
\authorrunning{Chang et al.}
%
\institute{Department of Biomedical Informatics \& Data Science, Yale University,\\
New Haven, United States \\
 \and
Department of Information Management, National Sun Yat-sen University, Kaohsiung, Taiwan\\
\faIcon{envelope}Corresponding authors:\email{shane.chang.tw@gmail.com},\email{syhwang@mis.nsysu.edu.tw}}

\maketitle              
\begin{abstract}

Topic modeling is widely used for uncovering thematic structures within text corpora, yet traditional models often struggle with specificity and coherence in domain-focused applications. Guided approaches, such as SeededLDA and CorEx, incorporate user-provided seed words to improve relevance but remain labor-intensive and static. Large language models (LLMs) offer potential for dynamic topic refinement and discovery, yet their application often incurs high API costs. To address these challenges, we propose the LLM-assisted Iterative Topic Augmentation framework (LITA), an LLM-assisted approach that integrates user-provided seeds with embedding-based clustering and iterative refinement. LITA identifies a small number of ambiguous documents and employs an LLM to reassign them to existing or new topics, minimizing API costs while enhancing topic quality. Experiments on two datasets across topic quality and clustering performance metrics demonstrate that LITA outperforms five baseline models, including LDA, SeededLDA, CorEx, BERTopic, and PromptTopic. Our work offers an efficient and adaptable framework for advancing topic modeling and text clustering.

\keywords{Topic Modeling \and Text Clustering \and Large Language Model}

\end{abstract}
\section{Introduction}

Topic modeling is a powerful method for uncovering thematic structures within large text corpora, enabling users to gain insights into the underlying topics of textual data. By clustering documents into topics based on word patterns, topic models support a wide range of applications, from exploratory data analysis to targeted decision-making in diverse fields like healthcare, social sciences, and business. Traditionally, unsupervised topic models such as Latent Dirichlet Allocation (LDA)~\cite{bleiLatentDirichletAllocation2003} have been employed for these tasks by learning topic distributions across words and documents, helping users to broadly explore the structure of a corpus.

While standard topic models can identify general themes, they often lack specificity and coherence in domain-focused applications. To address this, guided topic modeling approaches~\cite{jagarlamudiIncorporatingLexicalPriors2012,gallagherAnchoredCorrelationExplanation2017,mengDiscriminativeTopicMining2020,churchillGuidedTopicNoiseModel2022} allow users to introduce domain-specific topic seeds, improving the model’s alignment with expert knowledge and generating topics that are more coherent and more relevant to specific applications. For instance, models such as SeededLDA~\cite{jagarlamudiIncorporatingLexicalPriors2012}, CorEx~\cite{gallagherAnchoredCorrelationExplanation2017}, and GTM~\cite{churchillGuidedTopicNoiseModel2022} leverage seed word lists to help align the process of topic discovery, enhancing the interpretability and precision of topics based on the user’s goals. However, even with guided approaches, these models still face limitations in capturing nuanced relationships and require humans in the loop to guide the topic discovery, which is time-consuming and labor-intensive. 

With the advent of large language models (LLMs) like GPT and its successors, research on topic modeling has entered a new era. LLMs can leverage vast semantic understanding to enhance topic modeling by generating topic proposals, reassigning uncertain topics, and even discovering novel topics that conventional models might overlook. Despite these advantages, existing LLM-assisted methods often face practical challenges. Many require significant API usage by enriching documents \cite{viswanathanLargeLanguageModels2024},  providing constraints \cite{viswanathanLargeLanguageModels2024,zhangClusterLLMLargeLanguage2023}, and digesting every document in the corpus \cite{wangPromptingLargeLanguage2023,phamTopicGPTPromptbasedTopic2024}, leading to high computational costs and inefficiencies that make them impractical for large datasets. 

To address these challenges, we propose an \textbf{L}LM-assisted \textbf{I}terative \textbf{To}pic \textbf{A}ugmentation framework (LITA)\footnote{Our codes are available at \url{https://github.com/Text-Analytics-and-Retrieval/LITA-LLM-Iterative-Topic-Augmentation}.}, designed to balance LLM utility with cost-effectiveness. Our framework begins with user-provided topic seeds and employs an embedding model to cluster documents, followed by an iterative refinement and augmentation process, making it distinct from conventional methods. Specifically, LITA identifies ambiguous documents based on their proximity to cluster centroids and employ an LLM to assign only these documents to existing or new topics. By reassigning and dynamically augmenting new topics as needed, LITA achieves high-quality, coherent topics while minimizing LLM queries, enabling more scalable and adaptable topic modeling.

Our work presents three key contributions to the field of topic modeling:

\begin{itemize}
    \item \textbf{Efficient LLM-assisted Topic Modeling}: We introduce LITA, a framework that strategically incorporates LLM feedback only when necessary, drastically reducing API costs while leveraging the LLM’s ability to refine and discover nuanced topics.
    \item \textbf{LLM-assisted Iterative Refinement and Augmentation for Dynamic Topic Discovery}: We investigate the capabilities of LLMs as evaluators for topic assignments, using them as a signal for refining clusters and discovering new topics, which further enhances the framework’s adaptability and topic coherence over iterative rounds.
    \item \textbf{Empirical Validation Across Multiple Datasets and Metrics}: We validate LITA on two datasets, showing that it consistently outperforms five established models, including LDA \cite{bleiLatentDirichletAllocation2003}, SeededLDA \cite{jagarlamudiIncorporatingLexicalPriors2012}, CorEx \cite{gallagherAnchoredCorrelationExplanation2017}, BERTopic \cite{grootendorstBERTopicNeuralTopic2022}, and PromptTopic \cite{phamTopicGPTPromptbasedTopic2024}, across two topic quality and two clustering performance metrics.
\end{itemize}

In summary, our work bridges the gap between guided topic modeling and LLM-driven methods, offering an efficient, flexible framework that pushes the boundaries of topic discovery while managing the practical limitations of LLM usage.

\section{Related Works}

Topic modeling is a text clustering technique for uncovering thematic structures within a corpus. One notable foundational algorithm, LDA~\cite{bleiLatentDirichletAllocation2003}, infers document-topic and topic-word distributions, providing a framework for users to explore prevalent topics. Building on this, guided topic modeling allows users to inject domain knowledge into the topic identification process, tailoring results to specific needs and generating more coherent topics. SeededLDA~\cite{jagarlamudiIncorporatingLexicalPriors2012} is the pioneer that leverages seed word lists to help align the process of topic discovery. CorEx~\cite{gallagherAnchoredCorrelationExplanation2017} takes user-provided guidance as anchor words in the embedding space and learns maximally informative topics. CatE~\cite{mengDiscriminativeTopicMining2020} simultaneously learns the embedding for user-provided category names, documents, and words, resulting in a category-aligned topic embedding model. GTM~\cite{churchillGuidedTopicNoiseModel2022} modifies the LDA generative process to incorporate user-provided seed words, incrementally improving the seed words and identifying new topics. These models may face limitations in capturing nuanced relationships because they still learn topic structures from word co-occurrences without using the contextualized embedding captured by advanced language models. They also require humans in the loop to guide the topic discovery, which is time-consuming and labor-intensive. With the development of pre-trained language models such as BERT~\cite{vaswaniAttentionAllYou2017}, frameworks like BERTopic~\cite{grootendorstBERTopicNeuralTopic2022} have leveraged contextualized embeddings for topic clustering by applying traditional clustering algorithms, such as HDBSCAN or K-means, to these embeddings. Our proposed framework is inspired by BERTopic but incorporates the benefit from guided topic modeling approaches: it starts with human-provided topic seeds and then integrates the LLM’s text understanding capability, iteratively refining and augmenting clustering results based on LLM feedback.

Recent advances in large language models (LLMs) have driven 
prompt-based methods in topic modeling and text clustering \cite{viswanathanLargeLanguageModels2024,wangPromptingLargeLanguage2023,phamTopicGPTPromptbasedTopic2024,wangGoalDrivenExplainableClustering2023}. Viswanathan et al.~\cite{viswanathanLargeLanguageModels2024} propose using LLMs to enrich documents with key phrases and generate pair document constraints for better clustering. While both text enrichment and pairwise constraint generation lead to improved performances, they demand significant amount of LLM queries. PromptTopic~\cite{wangPromptingLargeLanguage2023} assigns topics by prompting LLMs with demonstrations to label each document, later merging similar labels either by word similarity or LLM-assisted comparison. TopicGPT~\cite{phamTopicGPTPromptbasedTopic2024} initiates topic modeling by generating a list of potential topics based on a sampled subset of documents and then uses the LLM to assign a topic to each document. GoalEx~\cite{wangGoalDrivenExplainableClustering2023} combines a multi-stage process of topic proposal, assignment, and selection, prompting LLMs to derive a list of explanations (as potential topics), followed by assigning texts based on these explanations and selecting a final topic for each document through integer linear programming. ClusterLLM~\cite{zhangClusterLLMLargeLanguage2023} uniquely incorporates LLM feedback to refine embeddings rather than clustering assignments, using ambiguous text triplets to enhance embedding quality. Despite their unique processes, these approaches request LLMs for every document, which makes them costly and impractical for large datasets. Their approaches also limit dynamic topic discovery as they rely on LLM-assigned labels without iterative refinement.

In contrast, our iterative topic augmentation framework balances efficiency and flexibility. Instead of requiring LLM queries for every document, it focuses on ambiguous documents only, thereby drastically reducing API usage. By integrating K-means with topic seed words, followed by LLM-driven reassignment and agglomerative clustering for the ambiguous documents, our framework enables novel topic discovery and iterative clustering refinement without incurring prohibitive costs.

\section{LLM-assisted Iterative Topic
Augmentation Framework}

\begin{figure}
    \centering
    \includegraphics[width=\linewidth]{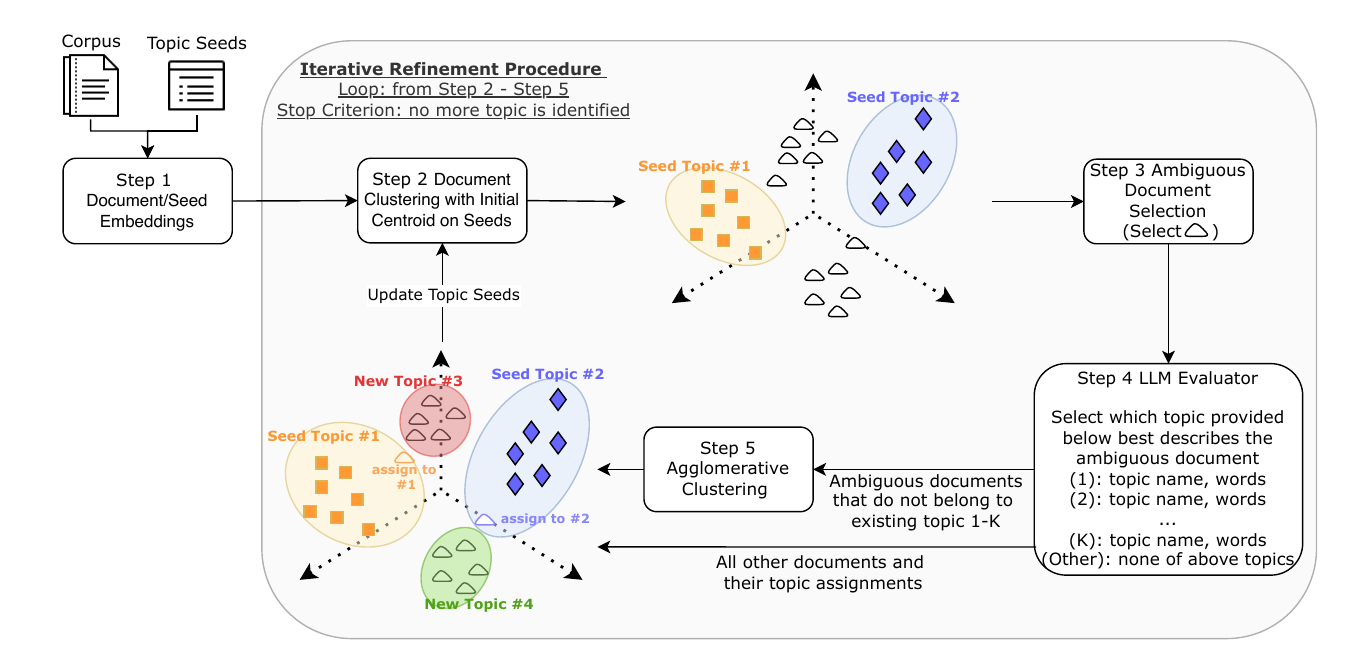}
    \caption{The Illustration of LITA}
    \label{fig:ITOREF}
\end{figure}

\subsection{Preliminary}

Our framework requires a corpus $\mathcal{D} = \{d_i\}^{N}_{i=1}$ and a set of user-provided seed word lists $\mathcal{S}=\{s_{i}\}^{|\mathcal{S}|}_{i=1}$, where $d_i$ represents a document, $N$ is the number of documents, and $s_i$ is a seed word list that represents a topic. The framework identifies $K$ topics from the corpus. Please note that users are only required to provide seed words for partial topics (i.e., $|\mathcal{S}| \leq K$). This framework iteratively refines topic assignments and augments new topics, providing flexibility in identifying the optimal number of topics $K$ based on the corpus.

Following the common topic modeling practice~\cite{bleiLatentDirichletAllocation2003,jagarlamudiIncorporatingLexicalPriors2012,gallagherAnchoredCorrelationExplanation2017,grootendorstBERTopicNeuralTopic2022}, each topic $k \in K$ is described by a set of representative words $\phi_k$. For a document cluster $k$, we identify these representative words of topic~$k$ using c-TF-IDF~\cite{grootendorstBERTopicNeuralTopic2022}:

\begin{equation}
    tf_{w,k} \times \text{log}(1+\frac{A}{tf_w}),
\end{equation}

\noindent where $tf_{w,k}$ is the word frequency of word $w$ in the cluster $k$, $A$ is the average word frequency across all clusters, and $tf_w$ is the word frequency of $w$ in the corpus. We select top-$M$ words with highest values of each topic to form $\phi_k$. We prompt an LLM to review represented words of each topic and generate an appropriate topic name.

\subsection{Procedure}

\subsubsection{Step 1, Document and Seed Word Embedding} In the first step, we embed all documents in the corpus and seed words using a pre-trained embedding model $f$. This yields document embeddings $\vec{\mathcal{D}} = \{\vec{d}_i = f(d_i) \}^{N}_{i=1}$ and seed word embeddings $\vec{\mathcal{S}} = \{\vec{s}_i = f(s_i) \}^{|\mathcal{S}|}_{i=1}$, placing them in the same vector space for direct comparison.

\subsubsection{Step 2, Document Clustering} We apply the K-means clustering to the document embeddings to identify topic (i.e., cluster) assignment of each document. To incorporate the user-provided seed words, we initialize the K-means with the seed word embeddings as the initial centroids, thereby injecting the initial topic structure based on user's guidance.

\begin{equation}
    \text{K-means}(\text{data}=\vec{\mathcal{D}}, \text{init} = \vec{\mathcal{S}}, \text{number of clusters} = |\mathcal{S}|)
\end{equation}

\subsubsection{Step 3, Ambiguous Document Identification}

Because users may provide only partial topic seed words, the initial clustering might not be able to fully capture the intended topic structure, leaving some documents ambiguously assigned. To address this, we identify ambiguous documents by evaluating their proximity to the two nearest cluster centroids. After K-means assigns each document to a cluster, we compute the centroid of each topic by averaging its member document embeddings. We then apply a margin-based metric to flag ambiguous documents:

\begin{equation}
    |\delta(\vec{d}_i, \mu_i^1) - \delta(\vec{d}_i, \mu_i^2)| \leq \epsilon,
\end{equation}

\noindent where $\delta$ is the cosine distance measuring the distance between two embeddings, and $\mu_i^1$ and $\mu_i^2$ are the nearest and second nearest topic centroids for $\vec{d}_i$. Ambiguous documents are those for which the distance difference is below the ambiguous distance threshold $\epsilon$, indicating uncertainty in their topic assignments.

\subsubsection{Step 4, LLM as Topic Evaluator}

Once ambiguous documents are identified, we leverage an LLM to evaluate topic assignments. An LLM is prompted to re-evaluate the topic assignment of each ambiguous document. To achieve this task, we create a prompt, which includes the task description, a ambiguous document $d_i$ , a list of topic names and their corresponding representative words, i.e., $\phi_k$ for each $k$. Therefore, an LLM determines one of the following cases for each ambiguous document:

\begin{itemize}
    \item (a) The document belongs to its assigned topic,
    \item (b) The document fits another existing topic, or
    \item (c) The document does not belong to any existing topics.
\end{itemize}

We retain and update the topic assignment for documents in cases (a) and (b), respectively. For those ambiguous documents in the case (c), we proceed to Step 5 to generate new topics. Our prompt template is as follows.

\begin{tcolorbox}[]

\verb|"""|

You are provided with the a list of topics and an article:

\vspace{\baselineskip}

Topics: \{topic name 1: topic words $\phi_1$, ...\}

\vspace{\baselineskip}
Article: \{an ambiguous document $d_i$\}

\vspace{\baselineskip}
Please select a topic that better corresponds with the article. Note that you must not generate a topic which is not in the provided topic list.

\vspace{\baselineskip}
Please respond with the topic name without explanation. If none of the topics better corresponds with the article in terms of topic, you should respond with `None' without explanation.

\verb|"""|
\end{tcolorbox}

\subsubsection{Step 5, Agglomerative Clustering for Identifying Unknown Topics}

We adopt the agglomerative clustering on ambiguous documents that belong to case (c) to identify potential new topic clusters. The reasons of the choice of agglomerative clustering are (1) it is free from setting the number of topics, which is important because the optimal number is unknown, and (2) it is a bottom-up clustering approach, which avoids augmenting a new topic cluster with only a single document. We use the euclidean distance and ward linkage criterion for the agglomerative clustering. The only parameter is the agglomerative distance threshold $\gamma$, which is dataset-dependent and empirically decided. The sensitivity analysis is reported in Fig.~\ref{fig:sensitivity}.

After applying agglomerative clustering, clusters containing at least five documents are considered new topics, and documents within these clusters receive new topic assignments. For those isolated documents not clustered in new topics, we keep using their originally assigned topics. As a result, we have the latest topic assignments to all documents. We apply c-TF-IDF (Eq. 1) to generate representative word lists for all topics and treat them as updated seed sets $\mathcal{S}$ for next refinement round. The framework repeats Step 2 -- Step 5 iteratively until no new topics are identified.
 
\section{Experiment}

\subsection{Experimental Setup}

\subsubsection{Datasets.} We evaluate the proposed LITA and baselines on two datasets, namely 20 Newsgroups~\cite{lang1995newsweeder} and CLINC(D)~\cite{zhangClusterLLMLargeLanguage2023}. The 20 Newsgroups is widely used for evaluating text classification and information retrieval tasks. The dataset encompasses 11,314 training data and 7,532 test data distributed across 20 topics. We only use the test data in our experiments. The CLINC(D) is a refinement of the CLINC dataset~\cite{larson-etal-2019-evaluation}, which was originally designed for intent classification. In~\cite{zhangClusterLLMLargeLanguage2023}, the authors remove the out-of-scope user utterances and utilize the domain of utterances as labels for domain discovery. The CLINC(D) dataset is stratified into 10 domains, each comprising 450 user utterances.

\subsubsection{Baselines.} 
We examine five widely used approaches, each based on distinct paradigms: LDA~\cite{bleiLatentDirichletAllocation2003}, BERTopic~\cite{grootendorstBERTopicNeuralTopic2022}, PromptTopic~\cite{wangPromptingLargeLanguage2023}, SeededLDA~\cite{jagarlamudiIncorporatingLexicalPriors2012}, and Anchored CorEX~\cite{gallagherAnchoredCorrelationExplanation2017}. LDA, the most traditional approach, is a probabilistic model that uncovers topic distributions based on word co-occurrences. BERTopic, a cluster-based method, leverages embeddings from pre-trained language models to create dense clusters and interpretable topic descriptions. PromptTopic, a recent LLM-based approach, contains three main steps: Topic Generation, Topic Collapse, and Topic Representation Generation. To mitigate the challenge of overlapping topics frequently generated by LLMs, the authors introduce two distinct methods for managing topic collapse: Prompt-Based Matching (PBM) and Word Similarity Matching (WSM). Considering a better performance of WSM in the original paper, WSM is used in the following experiments. Additionally, seed word guidance forms the basis of LITA, as well as SeededLDA and Anchored CorEx. SeededLDA extends LDA by biasing topic generation toward seed words.  Anchored CorEx also employs seed words to enhance interpretability but does not rely on generative assumptions. In our experiments, we maintain the default settings for most baselines, except for SeededLDA, for which we set the number of iterations to 100 and the refresh rate to 20. Additionally, we evaluate performance across different numbers of topics.

\subsubsection{Metrics.}
To assess the performance of the topic modeling methods, we employ four evaluation metrics to provide a comprehensive evaluation of both topic quality and clustering performance. Normalized Pointwise Mutual Information (NPMI)~\cite{bouma2009normalized} measures the semantic coherence of topics, ranging from -1 to 1, where higher values indicate better topic coherence. Topic Diversity (TD)~\cite{dieng2020topic} evaluates the distinctiveness of topics by calculating the ratio of unique terms across all topics, with higher scores reflecting greater diversity. On the other hand, we follow previous works~\cite{zhangClusterLLMLargeLanguage2023,viswanathanLargeLanguageModels2024} to evaluate clustering performance with Normalized Mutual Information (NMI)~\cite{strehl2002cluster} and clustering accuracy. To compute accuracy, we acknowledge that clustering algorithms assign labels arbitrarily. Therefore, we use the Hungarian algorithm~\cite{kuhn1955hungarian} to optimize the alignment between predicted and true labels, ensuring an accurate comparison.

\subsubsection{Implementation Details.} 

We adopt BGE-M3 model~\cite{chen2024bgem3embeddingmultilingualmultifunctionality} as the pre-trained embedding model. We use the OpenAI gpt-3.5-turbo-0125 API\footnote{API Available at: https://platform.openai.com/docs/models\#gpt-3-5-turbo.} as the backbone of the LLM Evaluator, and we set the temperature as 0 for reducing the randomness. Clustering methods are conducted using Scikit-Learn's~\footnote{https://scikit-learn.org} MiniBatchKMeans and AgglomerativeClustering implementations for Step 2 and Step 5, respectively. The ambiguous distance threshold $\epsilon$ is set to 0.1, while the agglomerative distance threshold $\gamma$ is configured to 1.1. A sensitivity analysis of both thresholds is presented in subsequent experiments.

\subsection{Results \& Analysis}

\begin{table}[]
\caption{Main Results: we report LITA's performance across iterations (i.e., 1, 3, 5, and 7 on 20Newsgroups and from 1 to 5 on CLINC(D) dataset). Although other baselines do not implement an iterative scheme, we report their performances on different numbers of topics ($K$) that match the number of topics generated by LITA in different iterations to ensure a fair comparison.}
\label{tab:results}
\resizebox{\textwidth}{!}{%
\begin{tabular}{@{}lllllllllll@{}}
\toprule
       &             & \multicolumn{4}{l}{20Newsgroups}          & \multicolumn{5}{l}{CLINC(D)}                               \\ \cmidrule(l){3-11} 
Metric & Method      & 1(5)           & 3(19)  & 5(25)  & 7(28)  & 1(3)           & 2(8)           & 3(11)  & 4(13)  & 5(14)  \\ \midrule
\multirow{6}{*}{NPMI} &
  LITA &
  0.121 &
  \textbf{0.277} &
  \textbf{0.279} &
  \textbf{0.304} &
  0.132 &
  0.264 &
  0.280 &
  \textbf{0.312} &
  \textbf{0.313} \\
 &
  PromptTopic &
  \textbf{0.126} &
  0.218 &
  0.233 &
  0.254 &
  0.135 &
  \textbf{0.279} &
  \textbf{0.289} &
  0.292 &
  0.301 \\
       & BERTopic    & 0.120          & 0.226  & 0.244  & 0.259  & 0.126          & 0.224          & 0.249  & 0.268  & 0.277  \\
       & corEx       & 0.111          & 0.218  & 0.249  & 0.250  & \textbf{0.139} & 0.243          & 0.272  & 0.276  & 0.282  \\
       & SeededLDA   & -0.129         & -0.122 & -0.112 & -0.109 & -0.131         & -0.157         & -0.152 & -0.152 & -0.149 \\
       & LDA         & 0.112          & 0.198  & 0.217  & 0.231  & 0.117          & 0.224          & 0.226  & 0.240  & 0.243  \\ \midrule
\multirow{6}{*}{Topic Diversity} &
  LITA &
  0.481 &
  0.548 &
  0.542 &
  0.559 &
  \textbf{0.701} &
  0.665 &
  \textbf{0.678} &
  \textbf{0.676} &
  \textbf{0.689} \\
       & PromptTopic & 0.521          & 0.527  & 0.540  & 0.536  & 0.672          & \textbf{0.675} & 0.661  & 0.649  & 0.655  \\
       & BERTopic    & 0.249          & 0.309  & 0.328  & 0.329  & 0.526          & 0.498          & 0.514  & 0.528  & 0.532  \\
       & corEx       & 0.431          & 0.425  & 0.433  & 0.445  & 0.581          & 0.552          & 0.549  & 0.542  & 0.536  \\
 &
  SeededLDA &
  \textbf{0.840} &
  \textbf{0.622} &
  \textbf{0.633} &
  \textbf{0.661} &
  0.643 &
  0.561 &
  0.583 &
  0.579 &
  0.564 \\
       & LDA         & 0.239          & 0.246  & 0.251  & 0.252  & 0.344          & 0.336          & 0.293  & 0.299  & 0.298  \\ \midrule
\multirow{6}{*}{NMI} &
  LITA &
  0.457 &
  \textbf{0.635} &
  \textbf{0.635} &
  \textbf{0.637} &
  \textbf{0.393} &
  \textbf{0.471} &
  \textbf{0.531} &
  \textbf{0.594} &
  \textbf{0.616} \\
       & PromptTopic & \textbf{0.459} & 0.630  & 0.632  & 0.633  & 0.388          & 0.467          & 0.521  & 0.581  & 0.612  \\
       & BERTopic    & 0.248          & 0.444  & 0.460  & 0.482  & 0.063          & 0.455          & 0.486  & 0.477  & 0.496  \\
       & corEx       & 0.055          & 0.184  & 0.214  & 0.222  & 0.105          & 0.257          & 0.262  & 0.288  & 0.258  \\
       & SeededLDA   & 0.371          & 0.479  & 0.434  & 0.454  & 0.056          & 0.132          & 0.174  & 0.206  & 0.164  \\
       & LDA         & 0.177          & 0.218  & 0.225  & 0.245  & 0.050          & 0.096          & 0.101  & 0.116  & 0.102  \\ \midrule
\multirow{6}{*}{Accuracy} &
  LITA &
  \textbf{0.241} &
  \textbf{0.560} &
  \textbf{0.549} &
  \textbf{0.547} &
  \textbf{0.288} &
  \textbf{0.516} &
  \textbf{0.550} &
  \textbf{0.616} &
  \textbf{0.618} \\
       & PromptTopic & 0.239          & 0.555  & 0.542  & 0.542  & 0.287          & 0.514          & 0.533  & 0.601  & 0.609  \\
       & BERTopic    & 0.145          & 0.320  & 0.367  & 0.368  & 0.140          & 0.382          & 0.389  & 0.396  & 0.392  \\
       & corEx       & 0.099          & 0.177  & 0.211  & 0.188  & 0.192          & 0.352          & 0.345  & 0.317  & 0.316  \\
       & SeededLDA   & 0.224          & 0.427  & 0.398  & 0.382  & 0.201          & 0.268          & 0.250  & 0.245  & 0.219  \\
       & LDA         & 0.135          & 0.228  & 0.229  & 0.241  & 0.168          & 0.230          & 0.206  & 0.238  & 0.186  \\ \bottomrule
\end{tabular}%
}
\end{table}

\subsubsection{Performances of Topic Modeling.}
Table~\ref{tab:results} presents the evaluation results in four metrics. In terms of topic coherence (measured by NPMI), LITA consistently demonstrates superior performance. LITA achieves the highest or second highest NPMI values in most cluster configurations. In comparison, PromptTopic, BERTopic, and CorEX exhibit similar but slightly lower coherence levels. Traditional methods such as LDA and SeededLDA perform substantially worse, whereas SeededLDA produces negative NPMI scores, highlighting its inability to generate semantically coherent topics. In addition to coherence, LITA demonstrates competitive performance in topic diversity. Maintaining high diversity while preserving coherence is a critical attribute for practical applications. Although SeededLDA records a maximum diversity score in the 20Newsgroups, its negative coherence metrics indicate a trade-off that undermines semantic quality. For clustering performance, the results reveal an upward trend in NMI and accuracy for LITA during iterations (as the number of clusters increases), reflecting improved alignment with the ground truth as the granularity becomes finer. Although PromptTopic achieves results comparable to LITA in NMI and accuracy, LITA’s enhanced efficiency offers a clear advantage, which will be explored in detail in the subsequent section.

\begin{figure}
  \centering
  \includegraphics[width=\columnwidth]{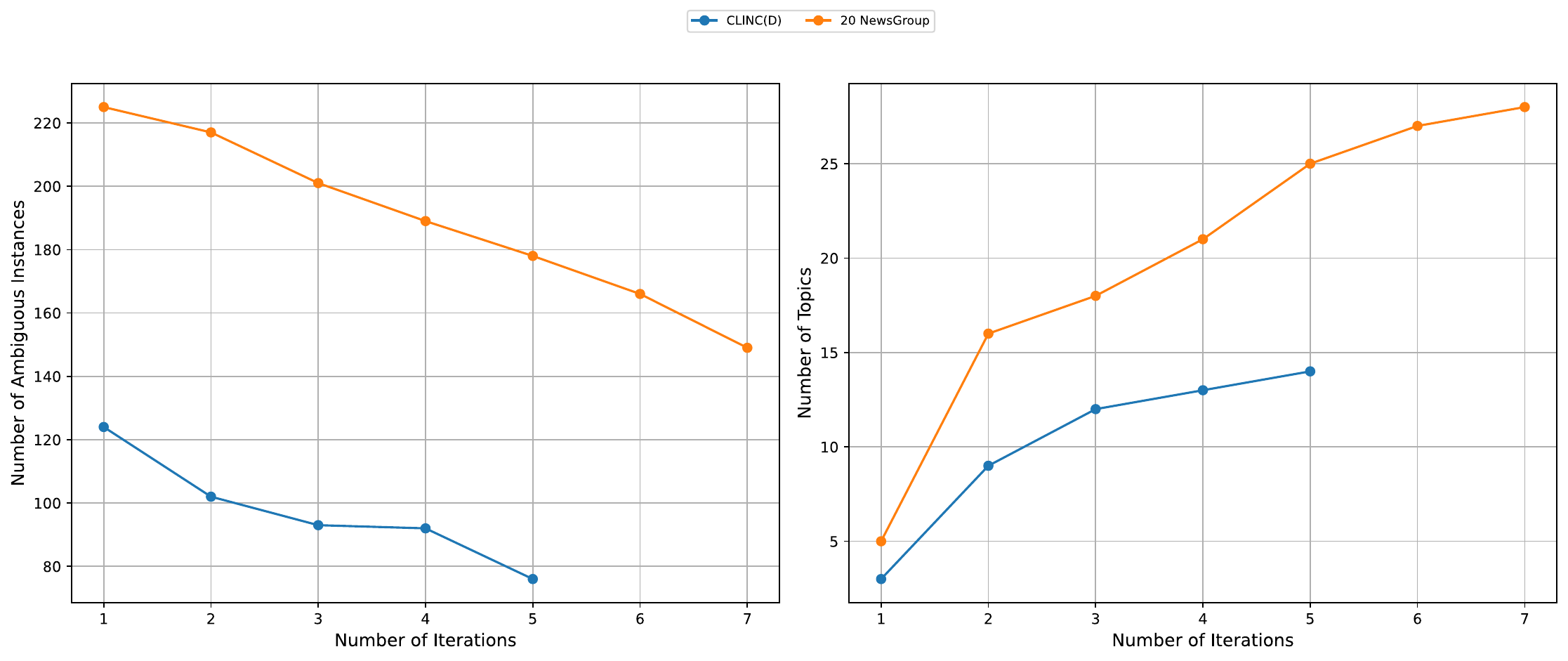}
  \caption{Number of Ambiguous Instances in each Iteration}
    \label{fig:iteration}
\end{figure}

\subsubsection{Efficiency Discussion.} Previous prompt-based methods necessitate a greater number of LLM requests due to their workflow. For instance, PromptTopic processes each document individually to assign topics, resulting in 7,532 and 4,500 API calls for the 20 Newsgroups and CLINC(D) datasets, respectively. Furthermore, PromptTopic requires additional API access during the topic collapse phase. In contrast, LITA strategically limits LLM invocations to only the evaluation of ambiguous instances. As illustrated in Figure~\ref{fig:iteration}, the number of ambiguous instances decreases substantially across iterations for both datasets, reaching topic convergence within just 5–7 iterations. As a result, LITA requires just 1,325 and 487 API calls on the same datasets. Compared to PromptTopic, LITA reduces over 80\% LLM requests on both 20 Newsgroups and CLINC(D) datasets. Considering to LITA's competitive performances reported in Table~\ref{tab:results}, LITA identifies topics of better quality while achieving computational efficiency, making LITA as cost-effective and scalable solution for real-world applications.

\subsubsection{Sensitivity Analysis.}
\begin{figure}{}
    \centering
    \includegraphics[width=\columnwidth]{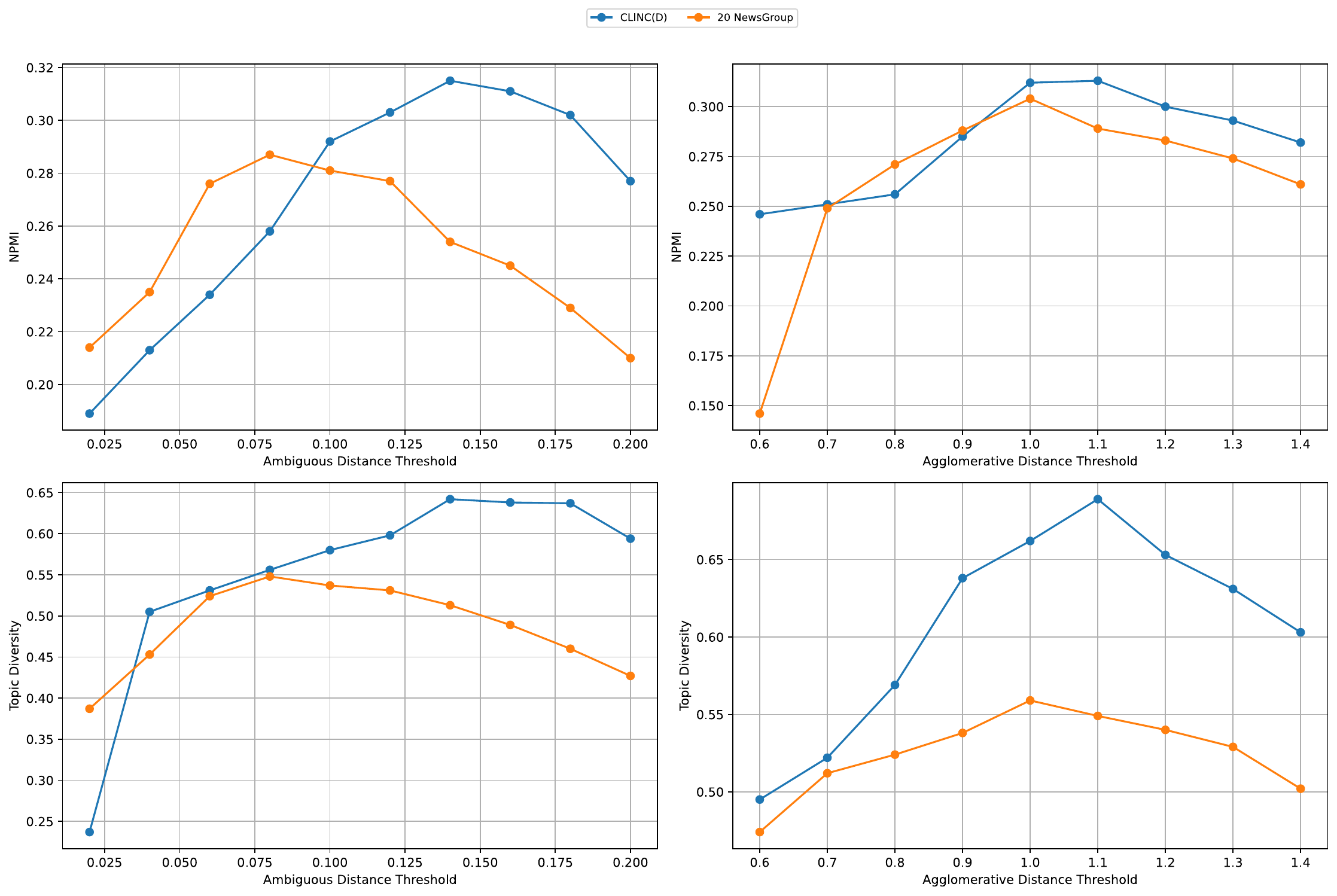}
    \caption{Sensitivity Analysis: we report the impact on the ambiguous distance threshold ($\epsilon$) and agglomerative distance threshold ($\gamma$) across both datasets.}
    \label{fig:sensitivity}
\end{figure}

Fig.~\ref{fig:sensitivity} illustrates that for the ambiguous distance threshold $\epsilon$, we observe an inverse-U shape effect in the NPMI and TD across two datasets, implying too loose (smaller $\epsilon$) or too strict (larger $\epsilon$) threshod degrades both NPMI and TD. For the agglomerative distance threshold $\gamma$, NPMI and TD initially improve with increasing $\gamma$ but decline as $\gamma$ continues to grow. Since $\gamma$ governs LITA to augment new topics, a large value of $\gamma$ can prevent new topic creation and cause the iterative process to terminate prematurely. Conversely, a smaller $\gamma$ leads to overly granular topics, negatively impacting performance.

\begin{table}{}
\centering
\caption{Ablation Study: we exclude Step 4 (LLM as the Evaluator) from LITA, allowing all ambiguous documents to proceed directly to the agglomerative clustering step. All numbers are the average performance from three different runs.}
\begin{tabular}{l|cc|cc}
\hline
\multirow{2}{*}{Method} & \multicolumn{2}{c|}{CLINC(D)} & \multicolumn{2}{c}{20NewsGroup} \\ \cline{2-5} 
                        & NPMI         & TD          & NPMI            & TD            \\ \hline
LITA               & 0.302        & 0.663       & 0.295               & 0.552             \\
w/o LLM Evaluator       & 0.244        & 0.521       & 0.251               & 0.513             \\ \hline
\end{tabular}
\label{tbl:abl}
\end{table}

\subsubsection{Ablation Study.}
Table~\ref{tbl:abl} highlights the importance of incorporating the LLM as Topic Evaluator in LITA. The results demonstrate the value of the LLM as Topic Evaluator in enhancing both topic coherence (NPMI) and topic diversity (TD). By leveraging LLMs to refine topic assignments, LITA is more effective in handling ambiguous documents, ensuring that topics are both coherent and diverse.

\section{Conclusion}
This paper presents LITA, an LLM-assisted iterative topic augmentation framework that balances cost-effectiveness and high-quality topic modeling. By leveraging LLMs to refine ambiguous documents and expand topics dynamically, LITA outperforms baseline methods across key metrics, including topic coherence, diversity, NMI, and clustering accuracy. Its efficient integration of LLM feedback without over-relying on costly API queries distinguishes it from other approaches. While LITA shows strong empirical results, limitations such as dependency on user-provided seeds and dataset-specific performance variations remain. Future work could focus on evaluating LITA on larger-scale and more domain-specific datasets to assess its generalizability and robustness. In conclusion, LITA provides a scalable and efficient solution for advancing topic modeling and demonstrates the potential of LLMs in this domain.

%
%
%
\bibliographystyle{splncs04}
\bibliography{refs}

\begin{thebibliography}{10}
\providecommand{\url}[1]{\texttt{#1}}
\providecommand{\urlprefix}{URL }
\providecommand{\doi}[1]{https://doi.org/#1}

\bibitem{bleiLatentDirichletAllocation2003}
Blei, D.M., Ng, A.Y., Jordan, M.I.: Latent dirichlet allocation. J. Mach. Learn. Res.  \textbf{3}(Jan),  993--1022 (2003)

\bibitem{bouma2009normalized}
Bouma, G.: Normalized (pointwise) mutual information in collocation extraction. Proceedings of GSCL  \textbf{30},  31--40 (2009)

\bibitem{churchillGuidedTopicNoiseModel2022}
Churchill, R., Singh, L., Ryan, R., {Davis-Kean}, P.: A {{Guided Topic-Noise Model}} for {{Short Texts}}. In: Proceedings of the {{ACM Web Conference}} 2022. pp. 2870--2878. ACM, Virtual Event, Lyon France (2022)

\bibitem{dieng2020topic}
Dieng, A.B., Ruiz, F.J., Blei, D.M.: Topic modeling in embedding spaces. Transactions of the Association for Computational Linguistics  \textbf{8},  439--453 (2020)

\bibitem{gallagherAnchoredCorrelationExplanation2017}
Gallagher, R.J., Reing, K., Kale, D., Ver~Steeg, G.: Anchored {{Correlation Explanation}}: {{Topic Modeling}} with {{Minimal Domain Knowledge}}. Transactions of the Association for Computational Linguistics  \textbf{5},  529--542 (2017)

\bibitem{grootendorstBERTopicNeuralTopic2022}
Grootendorst, M.: {{BERTopic}}: {{Neural}} topic modeling with a class-based {{TF-IDF}} procedure (2022), arXiv [cs.CL]

\bibitem{jagarlamudiIncorporatingLexicalPriors2012}
Jagarlamudi, J., Daum{\'e}, III, H., Udupa, R.: Incorporating {{Lexical Priors}} into {{Topic Models}}. In: Proceedings of the 13th {{Conference}} of the {{European Chapter}} of the {{Association}} for {{Computational Linguistics}}. pp. 204--213 (2012)

\bibitem{kuhn1955hungarian}
Kuhn, H.W.: The hungarian method for the assignment problem. Naval research logistics quarterly  \textbf{2}(1-2),  83--97 (1955)

\bibitem{lang1995newsweeder}
Lang, K.: Newsweeder: Learning to filter netnews. In: Machine learning proceedings 1995, pp. 331--339 (1995)

\bibitem{larson-etal-2019-evaluation}
Larson, S., Mahendran, A., Peper, J.J., Clarke, C., Lee, A., Hill, P., Kummerfeld, J.K., Leach, K., Laurenzano, M.A., Tang, L., Mars, J.: An evaluation dataset for intent classification and out-of-scope prediction. In: Proceedings of the 2019 Conference on Empirical Methods in Natural Language Processing and the 9th International Joint Conference on Natural Language Processing (EMNLP-IJCNLP) (2019)

\bibitem{mengDiscriminativeTopicMining2020}
Meng, Y., Huang, J., Wang, G., Wang, Z., Zhang, C., Zhang, Y., Han, J.: Discriminative {{Topic Mining}} via {{Category-Name Guided Text Embedding}}. In: Proceedings of {{The Web Conference}} 2020. pp. 2121--2132. ACM (2020)

\bibitem{phamTopicGPTPromptbasedTopic2024}
Pham, C.M., Hoyle, A., Sun, S., Iyyer, M.: {{TopicGPT}}: {{A Prompt-based Topic Modeling Framework}}. In: Proceedings of the 2024 {{Conference}} of the {{North American Chapter}} of the {{Association}} for {{Computational Linguistics}}: {{Human Language Technologies}}. vol.~1, pp. 2956--2984 (2024)

\bibitem{strehl2002cluster}
Strehl, A., Ghosh, J.: Cluster ensembles---a knowledge reuse framework for combining multiple partitions. Journal of machine learning research  \textbf{3}(Dec),  583--617 (2002)

\bibitem{vaswaniAttentionAllYou2017}
Vaswani, A., Shazeer, N., Parmar, N., Uszkoreit, J., Jones, L., Gomez, A.N., Kaiser, L., Polosukhin, I.: Attention is {{All}} you {{Need}}. In: Annual {{Conference}} on {{Neural Information Processing Systems}}. vol.~30, pp. 5998--6008 (2017)

\bibitem{viswanathanLargeLanguageModels2024}
Viswanathan, V., Gashteovski, K., Lawrence, C., Wu, T., Neubig, G.: Large {{Language Models Enable Few-Shot Clustering}}. Transactions of the Association for Computational Linguistics  \textbf{12},  321--333 (2024)

\bibitem{wangPromptingLargeLanguage2023}
Wang, H., Prakash, N., Hoang, N.K., Hee, M.S., Naseem, U., Lee, R.K.W.: Prompting {{Large Language Models}} for {{Topic Modeling}}. In: 2023 {{IEEE International Conference}} on {{Big Data}} ({{BigData}}). pp. 1236--1241 (2023)

\bibitem{wangGoalDrivenExplainableClustering2023}
Wang, Z., Shang, J., Zhong, R.: Goal-{{Driven Explainable Clustering}} via {{Language Descriptions}}. In: Proceedings of the 2023 {{Conference}} on {{Empirical Methods}} in {{Natural Language Processing}}. pp. 10626--10649 (2023)

\bibitem{zhangClusterLLMLargeLanguage2023}
Zhang, Y., Wang, Z., Shang, J.: {{ClusterLLM}}: {{Large Language Models}} as a {{Guide}} for {{Text Clustering}}. In: Proceedings of the 2023 {{Conference}} on {{Empirical Methods}} in {{Natural Language Processing}}. pp. 13903--13920 (2023)

\end{thebibliography}

\end{document}